%% file: main.tex
\documentclass[twocolumn]{article}

\usepackage{arxiv}

\usepackage[utf8]{inputenc} 
\usepackage[T1]{fontenc}    
\usepackage{hyperref}       
\usepackage{url}            
\usepackage{booktabs}       
\usepackage{amsfonts}       
\usepackage{nicefrac}       
\usepackage{microtype}      

\usepackage{graphicx}
\usepackage{natbib}
\usepackage{doi}
\usepackage{subfigure}
\usepackage{bbold}
\usepackage{amssymb}
\usepackage{mathabx}
\usepackage{subfiles}

\title{Biologically Inspired Semantic Lateral Connectivity for Convolutional Neural Networks}

\usepackage{authblk}

\author[1,3]{Tonio Weidler}
\author[2,4]{Julian Lehnen}
\author[2]{Quinton Denman}
\author[2]{Dávid Sebők}
\author[2]{Gerhard Weiss}
\author[2]{Kurt Driessens}
\author[1]{Mario Senden}

\affil[1]{Department of Cognitive Neuroscience\\
	Maastricht University\\
	The Netherlands}
\affil[2]{Department of Data Science and Knowledge Engineering\\
	Maastricht University\\
	The Netherlands}
\affil[3]{\texttt{uni@tonioweidler.de}}
\affil[4]{\texttt{julianlehnen@gmail.com}}

\date{}

\hypersetup{
pdftitle={Biologically Inspired Semantic Lateral Connectivity for Convolutional Neural Networks},
pdfauthor={Tonio Weidler, Julian Lehnen, Quinton Denman, Dávid Sebők, Gerhard Weiss, Kurt Driessens, Mario Senden},
pdfkeywords={Deep Learning, Convolutional Neural Networks, Machine Learning, Lateral Inhibition, Computer Vision, Neuroscience},
}

\begin{document}

\twocolumn[{
    \maketitle
    \begin{abstract}
    Lateral connections play an important role for sensory processing in visual cortex by supporting discriminable neuronal responses even to highly similar features. In the present work, we show that establishing a biologically inspired Mexican hat lateral connectivity profile along the filter domain can significantly improve the classification accuracy of a variety of lightweight convolutional neural networks without the addition of trainable network parameters. Moreover, we demonstrate that it is possible to analytically determine the stationary distribution of modulated filter activations and thereby avoid using recurrence for modeling temporal dynamics. We furthermore reveal that the Mexican hat connectivity profile has the effect of ordering filters in a sequence resembling the topographic organization of feature selectivity in early visual cortex. In an ordered filter sequence, this profile then sharpens the filters' tuning curves. 
    
    \end{abstract}
}]%

\section{Introduction}
\subfile{Sections/Introduction}

\section{Related Work}
\subfile{Sections/Related}

\section{Modeling Lateral Connectivity in CNNs}
\subfile{Sections/Approach}

\section{Experiments \& Results}
\subfile{Sections/Experiments}

\section{Connectivity Profile Effects}
\label{sec:profile-impact}
\subfile{Sections/MexicanHatEffect}

\section{Discussion \& Conclusions}
\label{sec:discussion}
\subfile{Sections/Discussion}

\bibliographystyle{unsrtnat}
\bibliography{references}

\end{document}

%% file: Sections/Introduction.tex
Neurons in primary visual cortex (V1) respond selectively to specific orientations present in an input image \citep{hubel1959receptive}. These neurons are organized in slabs perpendicular to the cortical surface commonly referred to as orientation columns. Neurons within a single column selectively respond to the same orientation while neurons in different columns respond to different orientations \citep{hubel1968receptive, hubel1974sequence}. Additionally, columns are organized by the similarity of their preferred orientations \citep{hubel1974sequence}. This likely results from the lateral connectivity profile of the visual cortex. Biological neurons locally influence each other through excitation and inhibition. In V1, lateral connections together give rise to a Mexican-hat profile wherein nearby neurons excite each other while more remote neurons mutually inhibit each other \citep{somers1995emergent, kang2003mexican}. As a result, a strong response to a specific orientation in the corresponding column inhibits those columns that correspond to marginally different angles. Lateral connections thereby sharpen the cortical response to the stimulus and thus improve orientation selectivity \citep{crook1998evidence, Isaacson2011lateral}.

In this work, we hypothesize that biologically inspired \textit{semantic}\footnote{We term any connectivity across channels \textit{semantic} since it links neurons activated by filters capturing different patterns.} lateral connectivity (SemLC) can be a beneficial extension for convolutional neural networks (CNNs), currently the go-to neural network architecture for dealing with visual input. We conceptually describe SemLC in Section~\ref{sec:approach} and discuss several strategies for effectively incorporating it into the structure of a convolutional layer. In Section~\ref{sec:experiments} we present an experimental evaluation of these strategies and empirically demonstrate as a proof-of-concept that SemLC improves the image classification performance of four different lightweight networks, including the recently proposed Capsule network \citep{sabour2017capsules}. Importantly, SemLC provides an effective structural improvement without requiring additional network parameters and hence retains a network's lightness. Furthermore, in Section~\ref{sec:order-experiments} we show that the lateral connectivity profile guides the learning of convolutional filters such that an ordering reflecting the similarity of their respective feature preference emerges. This ordering allows SemLC to be maximally effective and is highly comparable to that typically observed in V1. By confirming the emergence of such structure in CNNs, we thus provide converging evidence that a Mexican-hat lateral connectivity profile may indeed underlie the semantic ordering of orientation columns observed in the brain. We analyse the benefits of the profile by contrasting it to two alternatives in Section \ref{sec:profile-impact} and end the paper with a discussion of the benefits of SemLC and detail their roots in Section \ref{sec:discussion}. 

%% file: Sections/Related.tex
Lateral interactions play an important role for sensory processing in visual cortex \citep{somers1995emergent} and their potential benefit for connectionist computational models had been recognized as early as 1989 when \citeauthor{foldiak1989adaptive} introduced lateral interactions in Hebbian networks to further decorrelate neurons for principal component analysis. More recently, moving averages such as the local response normalization (LRN; \citeauthor{krizhevsky2012alexnet}, \citeyear{krizhevsky2012alexnet}) employ interactions between neighboring filters in a convolutional neural network. However, these interactions employ a uniform connectivity profile that inhibits a neuron's activity in proportion to the total energy in its local neighborhood. In contrast, we introduce weighted connections between cells whose strength depend on the relative position of cells in the filter sequence. Furthermore, in contrast to LRN, this (wavelet-based) approach introduces not only inhibition but excitation as well.

Whereas we establish connectivity in the semantic domain (i.e. between filters), previous research also studied lateral connectivity along spatially adjacent neurons in CNNs. \citet{fernandes2013lateral} build networks, in which neurons are inhibited by the weighted normalized sum of activity in the neuron's receptive field. \citet{cao2018lateral} showed that combining lateral inhibition with top-down feedback gives promising results on visual attention and saliency detection. To that end, the gradient of a class probability w.r.t. a neuron's activity gives rise to the correlation between the class and the pattern represented by the neuron, constituting an attention signal. These gradients are refined by a competition between neurons, modeled as a lateral inhibition zone around them. Similar to our work, \citet{cao2018lateral} employ a biologically inspired Mexican hat wavelet but expand it to a two-dimensional surface and apply it spatially to the activation map, rather than semantically to the filters. \citet{caballero2003spatio, caballero2014color, mira2004knowledge} perform segmentation of moving objects in image sequences where spatial and temporal lateral interactions between pixels \textit{charge} or \textit{discharge} their neighbours within a surrounding window. \citet{mao2007recurrent} study lateral connections in recurrent neural networks to improve the activity contrast between neurons by setting up a competition yielding a single victorious neuron. Unlike us, they assume a uniform lateral connectivity profile.

Due to their layered network architecture, scattering wavelet transforms \citep{mallat2012group, sifre2012combined} exhibit opportunities for lateral connections similar to those in CNNs. In fact, one may already consider the averaging transform applied to every layers output a spatial lateral connectivity since every neuron is modulated by the activity in its spatial neighborhood. Similarly, the recursive application of some set of wavelets $\tilde{W}$ constitutes lateral connectivity to the same degree to which consecutive layers in a CNN may be considered to laterally connect neurons. However, depth-dependent lateral connectivity is an incomplete model of true lateral interaction due to its negligence of temporal, and hence transitive and reciprocal, dynamics. \citet{sifre2012combined, sifre2013rotation} organize outputs of the averaging transform in orbits individually comprised of a full set of rotations in $\tilde{W}$ and apply another wavelet transform along the orbit (i.e. semantically), followed by an average. Consequently, a neuron's activity is modulated by its rotational neighborhood. Unlike the semantic lateral connectivity introduced in this work, \citet{sifre2012combined, sifre2013rotation} use their approach to build rotation invariant representations rather than to improve the orientation tuning of edge detectors. 

\citet{kavukcuoglu2009learning} introduce a feature extraction system comprised of a convolutional layer followed by a topographic organization of channels over which a Gaussian kernel pools the filter responses semantically. The topography is a two-dimensional map scanned by a pooling window outputting the Gaussian-weighted $\mathbf{L^2}$ matrix norm. Unlike \citet{sifre2012combined}, they learn the weights of the convolutional layer but also aim for rotation invariance. Due to Gaussian connectivity, their mechanism gives rise to a semantic structure of filters in the two-dimensional pooling map where similar filters tend to be closer to each other. In contrast to the Mexican hat wavelet used in this work, theirs uses a Gaussian connectivity profile which exclusively serves to facilitate but does not suppress neighboring filters.

\citet{gregor2011structured} introduce inhibitory lateral connections to impose a penalty for simultaneous firing. Similar to \citet{kavukcuoglu2009learning}, they organise neurons to determine neighborhoods in two-dimensional maps but also experiment on tree-structures. However, for every pair of neurons in a neighborhood of such a structure, an inhibiting connection is established. Similar to the results of \citet{kavukcuoglu2009learning} this gives rise to a semantic filter organization of like-oriented neighborhoods. Unlike ours, and in direct contrast to \citet{kavukcuoglu2009learning}, their approach involves exclusively inhibitory semantic lateral connectivity. Moreover, it is not integrated into a full CNN.

%% file: Sections/Approach.tex
\label{sec:approach}
We approach the modeling of lateral connectivity in convolutional neural networks (CNN) by employing weighted interactions between feature maps. In a 2D convolutional layer, each filter slides spatially over the input tensor to produce a scalar representation of every visited receptive field. The collection of these representations form \textit{output channels}, and the combination of all filters' output channels form a third-order tensor of activations. The mechanism of lateral connectivity introduced in this work lets these channels mutually influence each other.

\subsection{Biologically Inspired Connectivity}
Lateral connections between neurons in visual cortex exhibit a Mexican-hat profile \citep{muller2005attentional, kang2003mexican}. For this reason, we initialize lateral connections using a damped \textit{Ricker wavelet} $\omega$ with its peak at the modulated channel, given by

\begin{equation}
\label{eq:ricker}
    \omega(x) = \frac{2\delta}{\sqrt{3\sigma}\pi^{\frac{1}{4}}} \left( 1 - \frac{x^2}{\sigma^2} \right) e^{-x^2 / {2\sigma^2}}.
\end{equation}

\begin{figure}[t!]
\begin{center}
\centerline{\includegraphics[width=\columnwidth]{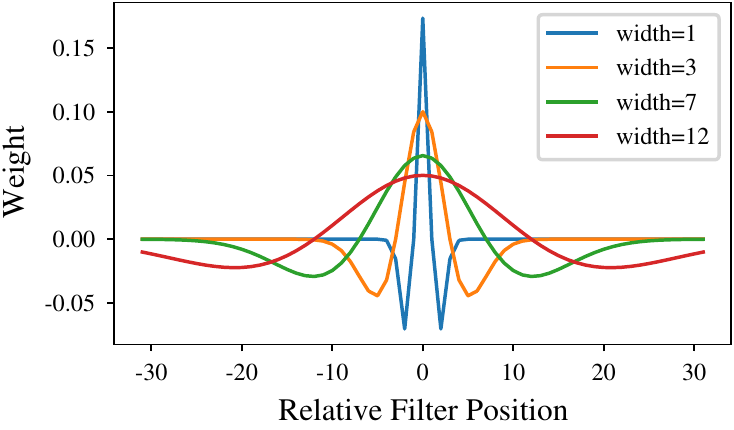}}
\caption{The Ricker wavelet formalizing the Mexican hat profile at different widths $\sigma$ (for its full function, see Equation \ref{eq:ricker}). As its peak is centered on the neuron at focus, this neuron's activity is facilitated by close neighbours but inhibited by remote cells.}
\label{fig:wavelet}
\vskip -0.2in
\end{center}
\end{figure}

The wavelet is parameterized by its width $\sigma$ and a damping factor $0 < \delta < 1$, which we introduce to avoid numerical instability by down scaling the wavelet. Figure \ref{fig:wavelet} shows examples for different values of $\sigma$. To construct the connectivity profile, the wavelet is discretized and centered around the modulated neuron. Note that we do not form a connection of a neuron to itself and therefore establish a weight of $0$ at the profile's center.

Following this connectivity profile, every channel is excited by close neighbors, but inhibited by filters at further distance. We hypothesize this to amplify regions in the filter sequence where multiple similar filters respond to the same input while increasing certainty in cases where the channels carry multiple hypotheses. A key requirement for this to work is a semantic filter ordering emerging during training. Note that the wavelet is approximately a Difference of Gaussians (DoG) and can hence also be regarded as a feature enhancement filter, which aligns with our hypothesis. 

\subsection{Temporal Dynamics}
We conceptualize the temporal evolution of the net input to a (fully connected) layer in an ANN as a dynamical system \citep[for a more detailed discussion of this refer to][]{yegnanarayana2009artificial}. In the case of constant input, such as image classification, and no recurrent connectivity, the temporal evolution of a layer's net input $\left(\mathbf{z}^l \right)$ is given by 
\begin{equation}
    \frac{d\mathbf{z}^l}{dt} = -\mathbf{z}^l + \mathbf{Wx}(t) + \mathbf{b},
\end{equation}
where $\mathbf{z}^l$ is a passive decay term, $\mathbf{x}(t)$ an external input modulated by the weight matrix $\mathbf{W}$ and $\mathbf{b}$ is a bias term. Finding the equilibrium of this system of independent differential equations (i.e. finding the converged values of $\mathbf{z}^l$) leads to the standard equation of the net input to a fully connected layer:
\begin{equation}
    \mathbf{z}^l = \mathbf{Wx} + \mathbf{b}.
\end{equation}

Introducing lateral connectivity renders the net input of units in a layer mutually dependent such that we obtain a coupled linear differential equation:
\begin{equation}
    \label{eq:diff-eq-lateral}
    \frac{d\mathbf{z}^l}{dt} = -\mathbf{z}^l + \mathbf{Uz}^l + \mathbf{Wx} + \mathbf{b},
\end{equation}
where $\mathbf{U}$ is the recurrent connectivity matrix. For the dynamics of such a system to be stable, its Jacobian ($ \mathbf{J} = \mathbf{U} - \mathbf{I}$, where $\mathbf{I}$ is the identity matrix) must have eigenvalues with a strictly negative real part; that is, the network feedback $\mathbf{U}$ must be sufficiently weak compared to local leakage (this is ensured by the $\delta$ parameter of the wavelet function).

Solving this system of coupled differential equations for the converged values of $\mathbf{z}^l$ one obtains
\begin{equation}
    \label{eq:diff-solved}
    \mathbf{z}^l = -\left(\mathbf{U} - \mathbf{I} \right)^{-1} \left( \mathbf{Wx} + \mathbf{b} \right).
\end{equation}
Simplifying this equation by introducing new terms $\mathbf{Q}=-\left(\mathbf{U} - \mathbf{I} \right)^{-1}$ and $\hat\mathbf{z}^l = \mathbf{Wx} + \mathbf{b}$, it is easy to see that the converged net input to a layer in the presence of linear feedback is equivalent to introducing an additional, linear, feed forward step:

\begin{equation}
    \label{eq:clc}
    \begin{array}{ll}
         \hat\mathbf{z}^l &= \mathbf{Wx} + \mathbf{b} \\
         \mathbf{z}^l &= \mathbf{Q}\hat\mathbf{z}^l
    \end{array}.
\end{equation}

\subsection{Algebraic Implementation}
\label{sec:lc-by-topelitz}
Since we realize lateral connectivity by introducing interactions between output channels of a convolutional layer, the output $\mathbf{Z}^l$ of the feed forward layer is a third-order tensor. In Equation \ref{eq:diff-eq-lateral}, the recurrent component $\mathbf{U}\mathbf{z}^l$ needs to be reformalized to apply to $\mathbf{Z}^l$. That is, a mathematical formulation of applying the connectivity profile to every neuron along the channel domain is required. Since lateral connectivity boils down to a moving weighted average it can in principle be realized by convolution. Specifically, since the connectivity profile ignores the spatial position of a neuron, the problem is a one-dimensional convolution of every depth-column with the wavelet. We implement this convolution in the form of a matrix multiplication with the Toeplitz matrix of the wavelet after unfolding the third-order input tensor. The resultant matrix product can then be folded back into the original input shape. An advantage of using a Toeplitz matrix is the ease of inverting the convolution, as required in Equation \ref{eq:diff-solved} to determine the equilibrium point. Alternatively one may conduct a deconvolution in the frequency domain by use of a fast Fourier transform (FFT). However, our benchmarking results demonstrate that this is substantially slower: a full forward and backward pass with Toeplitz-based SemLC is on average $4.55$ times faster than an FFT implementation. Generally, the added time cost of the former is approximately the same as adding a convolutional layer.

Note that using a Toeplitz matrix naturally renders the convolution circular, which aligns well with the organisation of cortical orientation columns. Laterally connecting kernels in a circle can also be motivated by the intuition that edge-detectors (as commonly emerge in the first layer of a CNN) when sorted by their orientation will form a ring because the last element of the sequence will again be similar to the first. However, choosing circular convolution over zero-padding is solely a conceptual decision, as we observed no significant difference between their influence on performance.

During training, the connectivity profile may either be fixed to its wavelet initialization or be allowed to adapt. Hereafter, we call the latter approach adaptive SemLC, or SemLC-A. Please note that an adaptive profile adds additional network parameters, while fixing it adds only (two) hyperparameters. Thirdly, we consider a variant in which the wavelet's parameters $\sigma$ and $\delta$ can be optimized (SemLC-P). Consequently, the optimization controls at which ranges and to which extent inhibition and excitation occur, but is constrained to the characteristic shape of the wavelet. This grants some adaptability while only adding two network parameters.

For the purpose of integrating lateral connectivity into a convolutional neural network, consider a classic pipeline of a convolutional block (convolution, ReLU, pooling). Semantic lateral connectivity will be added directly after the convolution because i) applying SemLC after pooling might result in relevant information being lost before the lateral connections can factor it in and ii) SemLC requires the dynamical system to be linear in order to analytically obtain the equilibrium.

%% file: Sections/Experiments.tex
\label{sec:experiments}
In the following, we experimentally examine the benefits of incorporating semantic lateral connectivity into different small-scale convolutional neural networks (CNN) and how it affects filter formation. As a proof of concept, we choose to limit our experiments towards such shallow architectures since the mechanism proposed here is based on inspiration from primary visual cortex, which we map to the first convolutional layer of a CNN.

\subsection{Proof of Concept}
\label{sec:strategy-comparison}

\begin{table*}[t!]
\caption{Comparison of SemLC variants to the baseline networks and Local Response Normalization (LRN) on CIFAR-10, with the classical split between train and test set. Confidence intervals are given at $\alpha = 0.05$.}
\label{tab:strategies}
\vskip 0.15in
\begin{center}
\begin{small}
\begin{sc}
\begin{tabular}{lrrr}
\toprule
Strategy            & AlexNet                           & Simple        & Shallow \\
\midrule
Baseline                & $84.91 \pm .07$           & $84.86 \pm .06$ & $79.52 \pm 0.17$ \\
LRN                 & $85.33 \pm .05$               & $85.38 \pm .06$ & $80.06 \pm 0.15$ \\
SemLC-G             & $85.21 \pm .05$               & $85.09 \pm .05$ & $79.36 \pm 0.23$ \\
\midrule
SemLC               & $85.52 \pm .06$             & $85.46 \pm .05$ & $81.00 \pm 0.11$ \\
SemLC-A             & $\mathbf{85.61} \pm .06$    & $\mathbf{85.65} \pm .06$ & $80.81 \pm 0.10$ \\
SemLC-P             & $85.48 \pm .06$             & $85.54 \pm .06$ & $\mathbf{81.66} \pm 0.19$ \\
\bottomrule
\end{tabular}
\end{sc}
\end{small}
\end{center}
\vskip -0.2in
\end{table*}

Our main set of experiments studies the effect of SemLC on three small-scale CNNs. We assess the different strategies in terms of the performance improvement their integration offers over the bare baseline networks on the classic image classification benchmark CIFAR-10 \footnote{\url{https://www.cs.toronto.edu/~kriz/cifar.html}} \citep{krizhevsky2009cifar}. The first network is a variant of AlexNet \citep{krizhevsky2012alexnet} built for the CIFAR-10 dataset. It comprises two convolutional and two locally connected layers, followed by a single, fully-connected softmax layer. We replace the locally connected layers with convolutional layers since the latter require substantially fewer parameters and add batch normalization \citep{ioffe2015batch} after the first two layers. The original AlexNet for CIFAR-10 is reported to achieve an error rate of $11\%$ on CIFAR-10, our reduced adaption achieves $15.09\%$. The second network, called Simple, is a variant of AlexNet with added dropout \citep{srivastava2014dropout} after the second and fourth convolutional layers and three fully connected layers in the classifier. The third architecture, Shallow, consists of only two convolutional blocks and a two-layer fully-connected classifier. By experimenting on three architectures we can test whether benefits from SemLC are exclusive to a specific network design.

With AlexNet, \citet{krizhevsky2012alexnet} also introduce local response normalization (LRN). This moving average normalizes filter responses to the same receptive field along the channel domain. As this resembles lateral connectivity but involves no connectivity profile we exclude it from the AlexNet baseline and add another baseline where the mechanism is included as a competitor to our mechanism. Additionally, we test SemLC with a Gaussian (SemLC-G) instead of a Mexican hat connectivity profile to control for the effect of the wavelet's shape.   

\paragraph{Hyperparameter Optimization} We determine a good (initial) width $\sigma$ and damping factor $\delta$ by randomly selecting $30$ unique random samples per strategy from the search space spanned by the Cartesian product of $\sigma \in \{3, 4, ..., 10\}$ and $\delta \in \{\frac{1}{10} + \frac{i}{70} | i \in \{1, 2, ..., 7\}\}$. For each random sample, the corresponding strategy is trained three times over $40$ epochs. A strategy's final set of hyperparameters is selected by averaging the best validation performances of the three independent runs. All experiments where conducted on the AlexNet adaption and then generalized to the other networks (which have the same number of filters in their first convolutional layer). We consistently (for all strategies) find $\sigma = 3$ and $\delta = 0.2$ to be the best configuration. 

\paragraph{Methodology} In all experiments, Adam \citep{kingma2015adam} serves as an optimizer with a learning rate threshold multiplied by $10$ once after $100$ and once after $150$ epochs. The training data is augmented by random cropping and horizontal flipping. Each strategy is evaluated by running $90$ independent training repetitions of $180$ epochs. From the training set we randomly sample $10\%$ of the data as a validation set. The final accuracy score of each model is the mean of the repetitions' best model accuracies on the test set. The best model is chosen based on validation accuracy. Confidence intervals are given based on the t-distribution with a $0.95$ percentile.

\subsubsection{Results}

Table \ref{tab:strategies} summarizes the results of the baseline, local response normalization (LRN), SemLC with a Gaussian profile (SemLC-G) and all three Mexican hat based SemLC strategies (SemLC, SemLC-A and SemLC-P) in the networks AlexNet, Simple and Shallow. LRN significantly improves performance of all networks \footnote{In the Shallow network, both the baseline and LRN experiments yielded outliers that substantially lower mean accuracy to $77.99 \pm 2.14$ and $77.72 \pm 2.65$ respectively. We removed them in the Table \ref{tab:strategies}'s data.}. At smaller margins, SemLC-G also significantly improves AlexNet and Simple, but has no significant effect on Shallow. Strikingly, all three Mexican hat based SemLC strategies consistently outperform the baseline models and both LRN and SemLC-G in all three networks. As part of AlexNet and Simple, the adaptive profile performs best, but both SemLC and SemLC-P are close runner-ups. Within the Shallow network, SemLC-P performs better than both alternative SemLC strategies.

\subsection{SemLC in Capsule Networks}
\begin{table}[ht!]
\caption{Added after the first (and sole) convolutional layer, SemLC is compared to LRN and the baseline in the standard CapsNet architecture on MNIST and CIFAR-10. Confidence intervals are given at $\alpha = 0.05$.}
\label{tab:capsnet}
\vskip 0.15in
\begin{center}
\begin{small}
\begin{sc}
\begin{tabular}{lrrr}
\toprule
Network             & MNIST                             & CIFAR-10         \\
\midrule
CapsNet             & $99.56 \pm .03$                   & $81.89 \pm .13$ \\
CapsNet + LRN       & $99.53 \pm .03$                   & $81.07 \pm .24$ \\
CapsNet + SemLC-G   & $99.54 \pm .02$                   & $80.57 \pm .25$   \\
\midrule
CapsNet + SemLC     & $99.56 \pm .02$                   & $82.14 \pm .22$ \\
CapsNet + SemLC-A   & $\textbf{99.58} \pm .02$                  & $81.40 \pm .25$ \\
CapsNet + SemLC-P   & $99.57 \pm .02$          & $\mathbf{82.18} \pm .19$ \\
\bottomrule
\end{tabular}
\end{sc}
\end{small}
\end{center}
\vskip -0.2in
\end{table}

To probe the performance of SemLC in a drastically different but nonetheless shallow architecture, we also incorporate it into the Capsule Networks (CapsNet) introduced by \citet{sabour2017capsules}. Just like \citet{sabour2017capsules}, we apply CapsNet to MNIST \footnote{\url{http://yann.lecun.com/exdb/mnist/}} \citep{lecun1998mnist} but also investigate performance on CIFAR-10 \citep{krizhevsky2009cifar}. CapsNet is a particular interesting case study for SemLC, because it consists of only a single convolutional layer, and as such relies heavily on the quality of its transformation. We optimize the hyperparameter $\sigma$ separately, probing $\sigma \in \{1, 3, 6, 12, 24, 32\}$ and again find $\sigma = 3$ to be the best wavelet width.

\subsubsection{Results}
Table \ref{tab:capsnet} summarizes this experiment. On MNIST, a baseline accuracy of $99.56\%$ already substantially limits possible margins of improvement. Neither LRN nor SemLC-G further improve this result and in fact both decrease accuracy by $0.03$ and $0.02$ percent points respectively. SemLC constitutes no change in accuracy, and SemLC-P and SemLC-A only numerically improve results. Applied to CIFAR-10, the situation is quite different. Whereas LRN and SemLC-G again significantly decrease CapsNet's performance, all other SemLC strategies improve the network by a small but significant margin. 

\subsection{Emerging Filter Organization}
\label{sec:order-experiments}

\begin{table}[t!]
\caption{Mean inter-filter mean squared deviation in the set of all filter pairs and the set of only adjacent filter pairs of different strategies and baselines. The percent decrease in MSD between the two sets indicates the order imposed by a strategy.}
\label{tab:mse}
\vskip 0.15in
\begin{center}
\begin{small}
\begin{sc}
\begin{tabular}{lrrr}
\toprule
Strategy             & All & Adjacent & \% Less Chaos\\
\midrule
None     & 0.231 & 0.230 &  0.4  \\
LRN    & 0.346 & 0.346  & 0.3  \\
SemLC-G & \textbf{0.413} & 0.264 & \textbf{36.1} \\
\midrule
SemLC     & 0.397 & 0.261  & 34.3  \\
SemLC-A   & 0.373 & 0.264  & 29.2  \\
SemLC-P & 0.397 & 0.256 & 35.5  \\
\bottomrule
\end{tabular}
\end{sc}
\end{small}
\end{center}
\vskip -0.1in
\end{table}

The LRN and SemLC networks trained in the previously described experiments all render filters dependent on their neighbors by introducing local connectivity. Naturally, such dependence should evoke an ordering in the filter sequence if the function of these lateral connections is not agnostic to the semantic relationship between the filter and its neighborhood. To study the emergence of such order, we quantify filter similarity by the mean squared deviation \footnote{Mean squared deviation refers to the same metric as mean squared error. We choose the different terminology to distinguish between the notions of an errors and differences.} (MSD) between their weights. Based on this metric, we first propose a numerical method of measuring induced order and then confirm the results in a qualitative analysis of the filter sequences. 

\subsubsection{Numerical Analysis}
\label{sec:numerical-analysis}

To detect order in a filter sequence we first measure the MSD of all $\frac{f(f-1)}{2}$ unique filter pairs in a sequence and then only the $f$ pairs of adjacent filters. Since our padding strategy supposes a circular ordering we also compare the last to the first filter. The proportional decrease from the mean all-pair MSD to the mean adjacent-pair MSD is then an indicator of order. Table \ref{tab:mse} reports these numbers. As expected, the base model exhibits no relevant difference between all-pair MSD and adjacent-pair MSD, reflecting a random ordering of the filter sequence. In contrast, all SemLC strategies show a clear MSD reduction, ranging from $29.2\%$ to $35.5\%$, where the parametric setting achieves the greatest reduction, but standard SemLC comes in as a close second at $34.3\%$. Notably, LRN induces no order despite locally connecting filters. SemLC-G on the other hand produces even higher order, with an MSD decrease of $36.1\%$. Observe, that additionally to the induced order, all SemLC variants and LRN constitute a higher overall inter-filter MSD. This hints towards an increased distinctiveness between the patterns they detect. 

\subsubsection{(Semi-)Qualitative Analysis}
\label{sec:qualitative-analysis}

\begin{figure}[t!]
\begin{center}
\centerline{\includegraphics[width=\columnwidth]{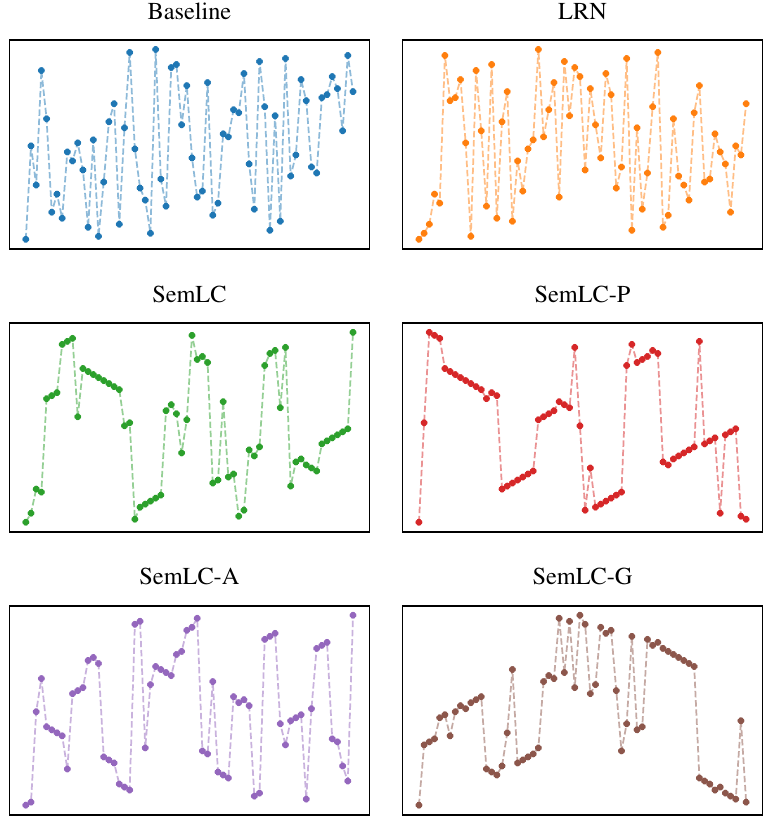}}
\caption{The original filter order compared to the order suggested by the 2-opt algorithm \citep{Croes58two-opt} with a threshold of $0.003$. The top two plots indicate no structure, but all filter sequences that emerged under some form of SemLC clearly exhibit subsequences of optimal order.}
\label{fig:ordering}
\end{center}
\vskip -0.2in
\end{figure}

\begin{figure*}[t!]
\begin{center}
\centerline{\includegraphics[width=\linewidth]{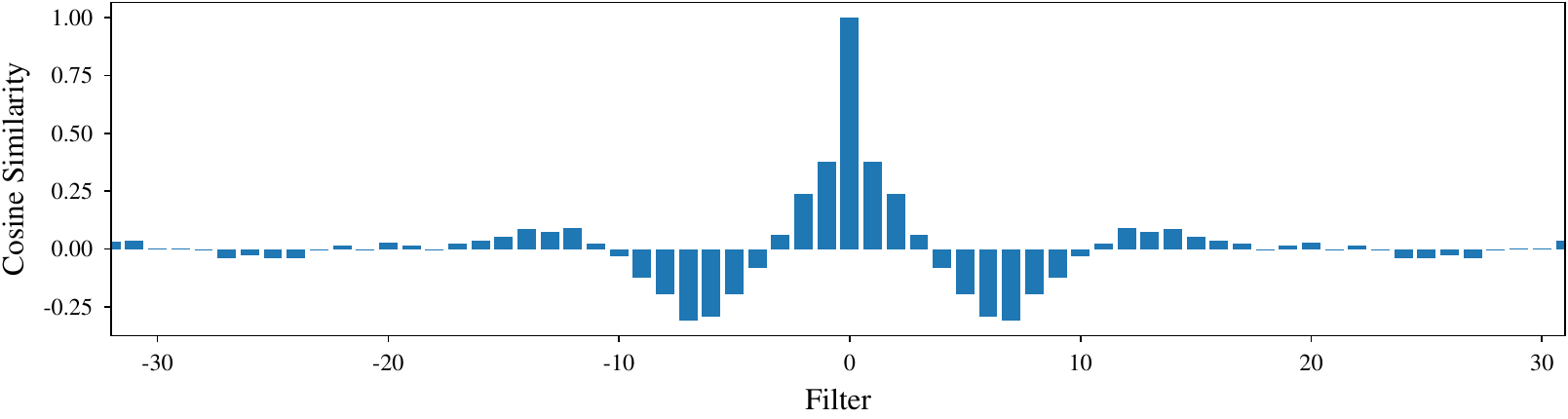}}
\caption{Average correlation of a filter to its neighbors in the full sequence, measured by cosine similarity. Each vector of cosine similarities (one per filter) is then shifted such that the filter based upon which all correlations where measured is always centered (placed at $0$). The filter sequence is that of an AlexNet with added SemLC trained on CIFAR-10.}
\label{fig:correlation}
\end{center}
\vskip -0.2in
\end{figure*}

To visualize the order in a filter sequence, we view the ordering of filters as a \textit{Traveling Salesman Problem} (TSP) where every filter is a vertex and edges are weighted by the inter-filter MSD. An optimal ordering is approximated by minimizing the total weight of a circular tour visiting all vertices. To this end, the 2-opt algorithm \cite{Croes58two-opt} can be applied to the network's order. We plot the result against the original ordering in Figure \ref{fig:ordering} by comparing indices. To better understand the meaning of such a plot consider two extremes: If the original ordering is random, the 2-opt algorithm will entirely restructure it and the plot will display randomly scattered points. If the original ordering is perfect, the two-opt algorithm will immediately stop and every filter will maintain its position in the sequence. The resultant plot is hence a diagonal line. All three SemLC variants impose some structure, but particularly those constrained to the wavelet exhibit well sorted subsequences. This aligns well with our findings from the numerical analysis.

The same conclusion can be drawn from Figure \ref{fig:correlation}. Here, we calculate the correlation of a filter to its neighbors based on their cosine similarity and average the results over all filters after centering the filter at focus. In an unordered sequence of filters, one expects a single peak in the center (since every filter maximally correlates to itself) and small correlation everywhere else (since randomly placed positive and negative correlation will level out through averaging). In Figure \ref{fig:correlation}, such a plot is created for an AlexNet with SemLC trained on CIFAR-10. It can be observed that filters have high positive correlation with their immediate two neighbors in both directions. This peak in positive correlation is flanked by a negative peak over about five filters on both sides, reflecting well the shape of the Mexican hat connectivity. The network is encouraged to place similar filters in the close neighborhood in which they mutually facilitate each others activity. Filters that correlate most negatively should instead be placed in the inhibitory zones of the connectivity profile. In consequence, their deactivation further facilitates the activation of the center neuron through the deactivation or even inversion of inhibitory connections. Filters around the range of the wavelet have low correlation, but group in shallow hills originating from similar filters co-occurring and thus conforming in their correlation with the center filter. 

%% file: Sections/MexicanHatEffect.tex
\begin{figure*}[t!]
\begin{center}
\centerline{\includegraphics[width=\linewidth]{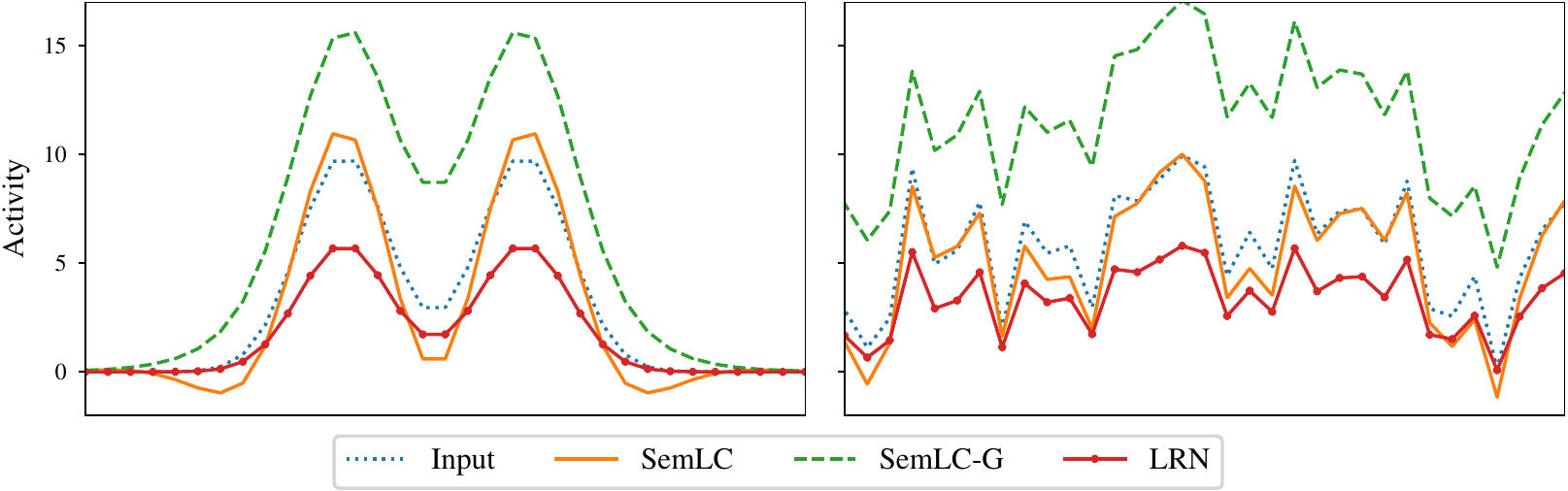}}
    \caption{Effect of local response normalization (LRN), SemLC with a Gaussian profile (SemLC-G) and SemLC with a Mexican hat profile, on two exemplary activity distributions. On the left, two bell-shaped activity peaks overlap. On the right, the activity distribution is generated uniform randomly.}
    \label{fig:lc-effects}
    \vskip -0.2pt
    \end{center}
\end{figure*}

We will now discuss how connectivity profiles differ in the effect they have on the activity distribution. Figure \ref{fig:lc-effects} depicts the effects of the Mexican hat profile (SemLC), the Gaussian profile (SemLC-G) and local response normalization (LRN; constituting a quasi-uniform profile) on two exemplary activity distributions. Applying a Gaussian profile generally smooths the distribution. As such, high-frequency information along the filter domain is suppressed. Cells with negative activity inhibit the activity of their neighbors - and positively active neurons facilitate their activation. Where there is all positive activation, the general level of activity rises and peaks and valleys converge. In its core neighborhood, a Mexican hat wavelet works similarly since it also conveys bell-shaped mutual excitation between close neighbors. However, inhibitory connections among neurons exhibiting further distance along the channel domain cause a competition in which high peaks will dominate regions of lower activity. Consequently, instead of smoothing the distribution, the Mexican hat will sharpen it. One can observe this particularly well on the left-hand side of Figure \ref{fig:lc-effects}. Here, the two overlapping peaks are elevated by their immediate neighborhood, but suppress the central valley and the surrounding tails, causing the peaks to separate again. Applied to the random activity distribution, one can observe low peaks being suppressed whilst high peaks sustain. In contrast, LRN normalizes the activity in local neighborhoods while also downscaling it based on a region's total energy. As such, it merges peaks and valleys but, unlike the Gaussian profile, also decreases the overall activity.

In the previous section, it was revealed that any type of Gaussian or Mexican hat profile will lead to an ordered filter sequence. The reason for this lies in the network's aim to reliably detect patterns (here mostly edges): to prevent short-range excitation from leaking activity into feature maps unrelated to the present signal the network is pressured to agglomerate filters responding to highly similar features. The result is semantic order. Since the Mexican hat wavelet adds medium-range inhibition, it additionally generates subsequences of similar feature detectors surrounded by clusters of filters preferring very different patterns. This is also reflected in the inter-filter correlation depicted in Figure \ref{fig:correlation}. Unlike SemLC, local response normalization fails to constitute order. We assume this to be caused by the purely inhibitory connectivity discouraging similar filters from placing next to each other since they would suppress each other. At the same time, it is impossible to surround every filter by only dissimilar filters, as they themselves would build homogeneous groups. In consequence, there is no incentive for the network to order the filter sequence. 

Lastly, an analysis of inter-filter mean squared deviations showed higher filter distinctiveness in networks with LRN and particularly SemLC. In the case of LRN, one can attribute this to the necessity of pattern detectors that remain distinctive after the smoothing normalization. The same rationale applies in light of the smoothing effect of a Gaussian profile. Under a Mexican hat profile, higher distinctiveness amplifies the sharpening effect, as a more homogeneous activity distribution might even cause inhibitory and excitatory influences to level each other out.

%% file: Sections/Discussion.tex
The present work introduces a novel type of lateral connectivity between channels to convolutional neural networks. We term this mechanism \textit{semantic} lateral connectivity (SemLC), because it interlinks neurons differing not in their respective receptive fields but in their feature tuning. Inspired by primary visual cortex, we let semantic lateral connectivity exhibit a Mexican hat shape. We show that it is possible to analytically solve for the activity profile resulting from recurrent interactions among feature channels enabling us to formulate these interactions as a pure feed forward computation. Specifically, we implement an exact and fast method by means of a Toeplitz matrix to solve for the equilibrium point of the linear dynamic system of neural activity. Experiments on image classification confirmed our two key hypotheses: a) employing interactions between channels following a Mexican-hat connectivity profile imposes similarity-based order on the sequence of filters and b) this order can in turn be leveraged by the lateral connections to improve the CNN's performance. In three small-scale convolutional neural networks, the proposed implementation of semantic lateral connectivity yielded significant improvements. Moreover, SemLC can boost the performance of the CapsNet \citep{sabour2017capsules} architecture, which utilizes just a single convolutional layer. Importantly, these performance gains are achievable without full profile adaptivity and thus by only introducing fixed connectivity needing no additional network parameters.

At the core of semantic lateral connectivity lies a Mexican hat wavelet, balancing mutual excitation between neighboring neurons and mutual inhibition between cells further apart. A key assumption of our biologically inspired \citep{muller2005attentional, kang2003mexican} approach is the benefit of this connectivity profile. Clear evidence towards this hypothesis provides the positive influence of our fixed (SemLC) and parametric (SemLC-P) approaches. Strikingly, allowing the network to deviate from the original wavelet had negligible benefit and even performed substantially worse than the fixed and parametric alternatives in two out of four networks. Furthermore, the same mechanism using a Gaussian profile (pure excitation) induces significantly smaller benefits. Local response normalization \citep[LRN;][]{krizhevsky2012alexnet} also causes significantly smaller improvements. In fact, unlike SemLC, the lateral interactions LRN models through a uniform inclusion of local activity do not benefit every architecture and even lower the performance of the Capsule Network. Together with the weaker performance gains of the Gaussian profile, this provides clear evidence for the utility of using a Mexican hat wavelet over other connectivity profiles. In fact, as we argued in the previous section, although both the Gaussian and the Mexican hat connectivity profile order the filter sequence, the added complexity of the latter leverages the emerging structure more efficiently and improves the layer's representation by sharpening its tuning curves.

To the deep learning community our results present an improvement to light-weight convolutional neural networks. The margin of improvement constituted by semantic lateral connectivity can certainly also be achieved by additional depth. However, a structural implementation of SemLC obviates the need to introduce and hence optimize additional network parameters. Our results further confirm the growing conviction that deep learning architectures can be enhanced by implementing principles and mechanisms derived from neuroscientific insights \citep[see e.g.][]{hassabis2017neuroscience}. Likewise, our findings provide insights relevant to neuroscientists interested in early visual cortex as we provide additional evidence that a Mexican hat connectivity profile may both guide the formation of orientation maps in the cortex and contribute to human object recognition by sharpening neurons' response profiles. Furthermore, our approach allows for the development of more biologically plausible CNNs in computational neuroscience, a field that increasingly utilizes deep learning in goal-driven research aimed to uncover representations and computations that may underlie complex functions of the biological brain \citep{yamins2016using}. Finally, the structure that SemLC imposes on the filter sequence can pave the way for further brain inspired extensions that benefit from topographic organization.